\documentclass[acmtog]{acmart}
\usepackage{multirow}
\usepackage{booktabs}
\usepackage{enumitem}
\copyrightyear{2025}
\acmYear{2025}
\makeatletter
\def\@ACM@copyright@check@cc{}
\makeatother
\setcopyright{cc}
\setcctype{by-nc}
\acmConference[SIGGRAPH Conference Papers '25]{Special Interest Group on Computer Graphics and Interactive Techniques Conference Conference Papers }{August 10--14, 2025}{Vancouver, BC, Canada}
\acmBooktitle{Special Interest Group on Computer Graphics and Interactive Techniques Conference Conference Papers (SIGGRAPH Conference Papers '25), August 10--14, 2025, Vancouver, BC, Canada}
\acmDOI{10.1145/3721238.3730614}
\acmISBN{979-8-4007-1540-2/2025/08}
\usepackage[ruled]{algorithm2e} %

\SetAlFnt{\small}
\SetAlCapFnt{\small}
\SetAlCapNameFnt{\small}
\SetAlCapHSkip{0pt}

\begin{document}
\title{GSHeadRelight: Fast Relightability for 3D Gaussian Head Synthesis}

\author{Henglei Lv}
\orcid{0009-0008-1907-0996}
\affiliation{%
 \institution{Institute of Computing Technology, CAS and University of Chinese Academy of Sciences}
 \streetaddress{6 Kexueyuan Nanlu}
 \city{Haidian Qu}
 \state{Beijing Shi}
 \postcode{100190}
 \country{China}
}
\email{lvhenglei22s@ict.ac.cn}

\author{Bailin Deng}
\affiliation{%
 \institution{Cardiff University}
 \city{Cardiff}
 \country{United Kingdom}
}
\email{DengB3@cardiff.ac.uk}

\author{Jianzhu Guo}
\affiliation{%
\institution{Kuaishou Technology}
 \city{Haidian Qu}
 \state{Beijing Shi}
\country{China}}
\email{guojianzhu1994@gmail.com}

\author{Xiaoqiang Liu}
\affiliation{%
 \institution{Kuaishou Technology}
 \city{Haidian Qu}
 \state{Beijing Shi}
 \country{China}
}
\email{liuxiaoqiang@kuaishou.com}

\author{Pengfei Wan}
\affiliation{%
 \institution{Kuaishou Technology}
 \city{Haidian Qu}
 \state{Beijing Shi}
 \country{China}
 }
\email{wanpengfei@kuaishou.com}

\author{Di Zhang}
\affiliation{%
 \institution{Kuaishou Technology}
 \city{Haidian Qu}
 \state{Beijing Shi}
 \country{China}
}
\email{zhangdi08@kuaishou.com}
\author{Lin Gao}
\authornote{Corresponding author is Lin Gao (gaolin@ict.ac.cn).}
\affiliation{%
 \institution{Institute of Computing Technology, CAS and University of Chinese Academy of Sciences}
 \streetaddress{6 Kexueyuan Nanlu}
 \city{Haidian Qu}
 \state{Beijing Shi}
 \postcode{100190}
 \country{China}
}
\email{gaolin@ict.ac.cn}

\renewcommand\shortauthors{Lv, H. et al}

\begin{abstract}
Relighting and novel view synthesis of human portraits are essential in applications such as portrait photography, virtual reality (VR), and augmented reality (AR). Despite recent progress, 3D-aware portrait relighting remains challenging due to the demands for photorealistic rendering, real-time performance, and generalization to unseen subjects. Existing works either rely on supervision from limited and expensive light stage captured data or produce suboptimal results. Moreover, many works are based on generative NeRFs, which suffer from poor 3D consistency and low real-time performance. We resort to recent progress on generative 3D Gaussians and design a lighting model based on a unified neural radiance transfer representation, which responds linearly to incident light.
Using only in-the-wild images, our method achieves state-of-the-art relighting results and a significantly faster rendering speed ($\times12$) compared to previous 3D-aware portrait relighting research. Project page: \url{http://www.geometrylearning.com/GSHeadRelight/}.

\end{abstract}

\ccsdesc[500]{Computing methodologies~Image manipulation}

\keywords{3D portrait relighting, 3D Gaussian splitting, radiance transfer}

\begin{teaserfigure}
  \includegraphics[width=\textwidth]{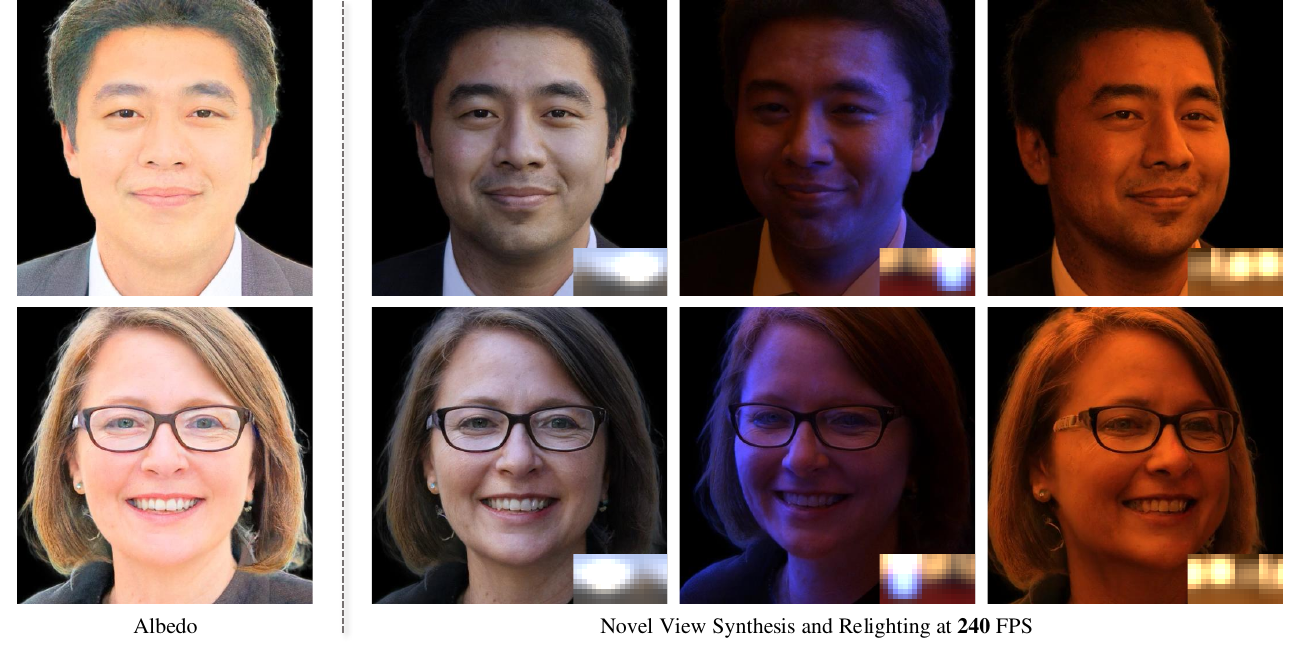}
  \caption{Our method incorporates an efficient yet effective lighting model into generative 3D Gaussian and can synthesize high-quality relightable 3D Gaussian heads that allow for novel view synthesis and relighting given any HDRI environment maps. Our method does not require expensive light stage data and achieves real-time rendering at 240 FPS, surpassing the previous 3D-aware portrait relighting research by at least 12 times.
  }
  \Description{This is the teaser figure for the article. }
  \label{fig:teaser}
\end{teaserfigure}

\maketitle

\section{Introduction}

Relighting and novel view synthesis of human portraits have always been two fundamental demands in various applications. In portrait photography, finding the ideal lighting and viewpoint often requires intensive expert knowledge and extensive trial and error. In virtual reality and augmented reality, virtual avatars need to be rendered under different lighting conditions and viewpoints in real-time and high fidelity. These demands drive the rapid development of the emerging field of 3D-aware portrait relighting. 

However, 3D-aware portrait relighting remains highly challenging due to its three demands: photorealistic appearance, real-time rendering, and one-shot generalization to unseen subjects. Human heads consist of highly intricate and varied materials that display different scattering and reflectance properties, making them among the most challenging objects to model accurately. The demand for real-time rendering significantly constrains algorithm design and model capabilities, and the requirement for sparse data input (single image or video) in many real-world applications brings the ambiguity between lighting and materials. Solving this under-constrained problem inevitably requires leveraging the powerful visual prior from pre-trained generative models. 

It has been proven that Generative Adversarial Networks (GANs) \cite{goodfellow2014generative} can provide such a prior. When incorporated with 3D representations, GANs can even learn 3D geometries from mere image datasets. Among them, generative NeRF~\cite{chan2022efficient, mildenhall2021nerf} find the balance between photorealism and 3D consistency.
The 3D head priors have enabled 3D-aware manipulation over human portraits such as attribute editing~\cite{gao2023sketchfacenerf, sun2022ide, jiang2025towards},  style transfer~\cite{yuan2024munerf} and relighting~\cite{jiang2023nerffacelighting, mei2024holo, rao2024lite2relight, cai2024real}. To resolve the ambiguity between lighting and materials, some works~\cite{mei2024holo, rao2024lite2relight} employ light stage data as ground truth for supervision. Nevertheless, light-stage captured data are expensive and usually closed source, so they are not accessible to common researchers. For better generalization and scalability, learning from casual image datasets is preferred. \citet{jiang2023nerffacelighting} distill albedo from weakly labeled image datasets at the cost of handling only white light. \citet{deng2024lumigan} propose an inverse rendering framework that models radiance transfer for the diffuse term and uses a simple BRDF for the specular term. Impressive as these works are,  most of them are based on generative NeRF, which is prone to limited rendering speed and 3D inconsistencies because of volume rendering and upsampling. More importantly, these works rely on neural rendering and need network evaluation for each lighting, which further limits the real-time performance of relighting.

We introduce GSHeadRelight, a relightable 3D head generative model with the latest generative 3D Gaussian backbone~\cite{hyun2024adversarial, kirschstein2024gghead}. Inspired by precomputed radiance transfer~\cite{sloan2002precomputed}, we design a lighting model based on Unified Neural Radiance Transfer. We assign spherical harmonics coefficients for each 3D Gaussian ellipsoid to model the radiance transfer, which handles visibility and global illumination, including self-occlusion and subsurface scattering. By making radiance transfer coefficients a function of view directions, our method can model both diffuse terms and low-frequency reflections. However, without explicit supervision of albedo, the ambiguity between albedo and lighting severely affects relighting quality. Since most in-the-wild images are under white illumination, we make a simple white-light assumption during training to mitigate the ambiguity, which provides plausible albedo decoupling. One desirable feature of radiance transfer is that the outgoing radiance, i.e. irradiance, responds to incident light linearly. Although our model is trained solely on white light, it can generalize well to colored illumination during test time. Moreover, since all components are organized as Gaussian attributes independent of lighting, only one forward inference is needed for head synthesis, and relighting can be performed without further generator evaluations. As a result, our method is extremely efficient and significantly reduces hardware requirements.

To summarize, we present three main contributions:
\begin{itemize}[leftmargin=*,topsep=1pt]
\item A relightable 3D Gaussian head generative model based on unified radiance transfer that supports global illumination.
\item A regime that distills plausible albedo and light transport functions from in-the-wild images without a heavy light stage setup.
\item Extensive experiments show that our method achieves state-of-the-art 3D-aware relighting results with extreme efficiency. Our model achieves a rendering speed of \textasciitilde 240 FPS on a single NVIDIA H800, with at least $12\times$ speedup over previous methods.

\end{itemize}

\begin{figure*}[ht]
    \centering
    \includegraphics[width=0.8\linewidth]{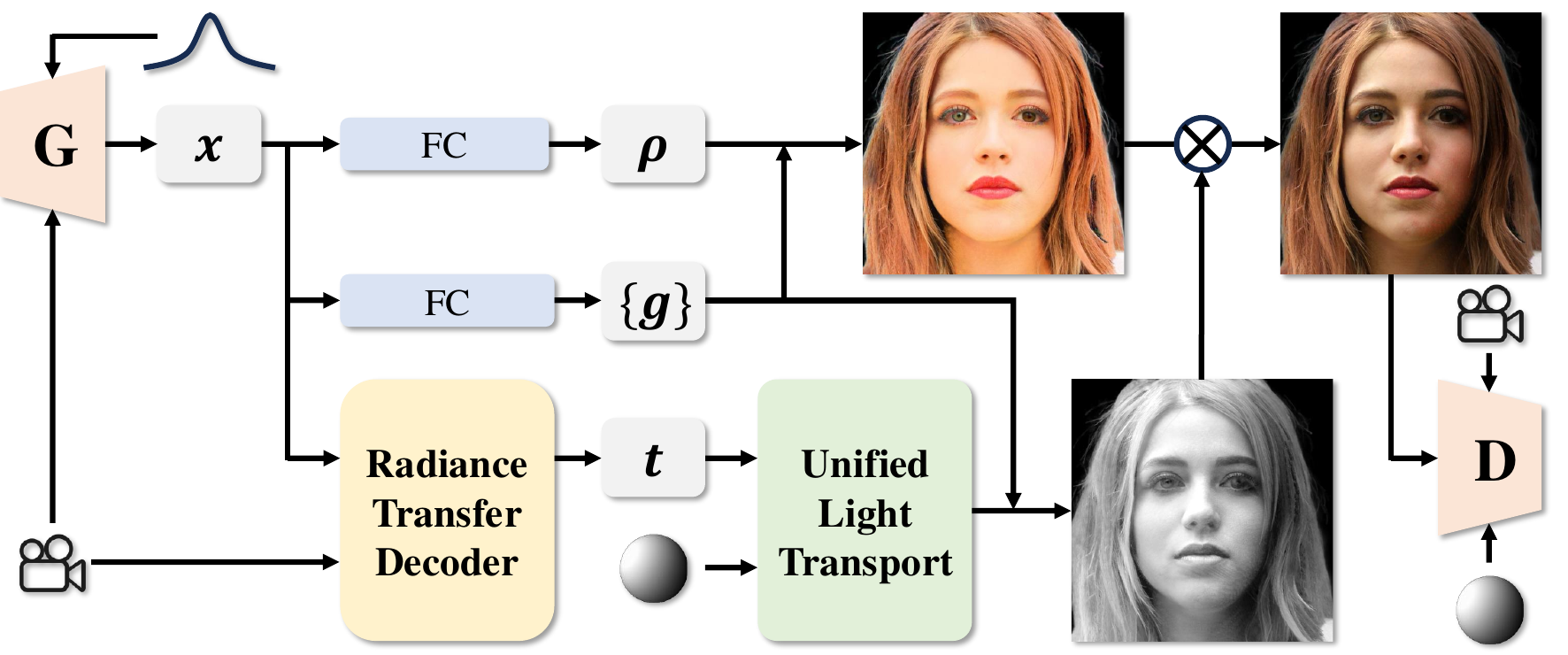}
    \caption{We incorporate a lighting model based on unified radiance transfer into a generative 3D Gaussian framework. The generator G takes in Gaussian noise and camera pose as condition and generates for each Gaussian an embedding $\mathbf{x}$, which is then linearly transformed into the albedo $\mathbf{\rho}$ and geometry attributes $\{\mathbf{g}\}$ including position, scale, rotation and opacity. A light-weight decoder conditioned on view direction transforms $\mathbf{x}$ into radiance transfer coefficients $\mathbf{t}$, which are used to compute the unified light transport with the light condition. Color is obtained by multiplying the light transport component with the albedo. The image is then rendered by standard splatting and sent to the discriminator D, which takes both camera pose and light condition as input. Mapping modules are omitted for simplicity. Gaussian noise and camera pose are represented as icons, and light condition as a diffuse sphere. 
   }
    \label{fig::pipeline}
    
\end{figure*}

\section{Related Work}

In this section, we review relevant work in the field of 3D-aware portrait synthesis.  We then examine deep neural face relighting from two perspectives: studio-captured data and casual image datasets.

\paragraph{3D-aware Portrait Synthesis}
Despite Diffusion Models (DMs)~\cite{ho2020denoising} being proven to have stronger model capabilities than GANs~\cite{goodfellow2014generative} in synthesizing images~\cite{rombach2022high}, videos~\cite{videoworldsimulators2024} and 3D assets~\cite{zhang2024clay}, GANs excel at 3D-aware portrait synthesis. The sparsity of high-quality 3D head models hinders the training of DMs, whereas 3D-GANs can learn fine geometry from mere image datasets.

Early 3D-GANs directly render pixels from explicit or implicit 3D representations, such as meshes~\cite{chen2019learning, henderson2020leveraging, kanazawa2018learning} and voxels~\cite{gadelha20173d, henzler2019escaping}, or implicit neural radiance fields~\cite{chan2021pi, schwarz2020graf}. Some works~\cite{nguyen2019hologan, niemeyer2021giraffe, xue2022giraffe} append neural rendering on the image side to improve image fidelity and enable high-resolution rendering. Subsequent works~\cite{gu2021stylenerf, or2022stylesdf, chan2022efficient} mostly adopt StyleGAN~\cite{karras2019style} as the generator due to its success. Among them, EG3D~\cite{chan2022efficient} proposes a hybrid radiance field representation and applies a super-resolution module to the generated tri-plane features. Despite its high expressiveness, volume rendering is too slow for native high-resolution synthesis, and super-resolution inevitably causes spatial inconsistencies across views. 

Recent works~\cite{hyun2024adversarial, kirschstein2024gghead} utilize 3D Gaussian Splatting (3DGS)~\cite{kerbl20233d} as the new explicit representation, allowing for native high-resolution head synthesis in real-time. \citet{kirschstein2024gghead} bind 3D Gaussians to a template mesh and generate UV maps of each Gaussian attribute under the StyleGAN~\cite{karras2019style} backbone. Gaussian ellipsoids are then queried and mapped back onto the template at arbitrary resolution. \citet{hyun2024adversarial} instead explore a hierarchical generative paradigm without a template mesh, where at each layer 3D Gaussians are constrained by anchor points from the previous layer. A new generative architecture is designed to generate multiple-level anchors and Gaussians. Our method is based on the recent progress in generative 3D Gaussian head synthesis. 
We refer to~\cite{sun2024recent} for more recent advances in 3D Gaussian Splatting.

\paragraph{Deep Neural Face Relighting}
Portrait relighting requires decomposition of appearance, geometry, and lighting. To acquire accurate facial reflectance fields, a line of works resort to meticulously calibrated studio equipment, including the widely adopted light stage~\cite{debevec2000acquiring}.  Neural networks are incorporated into the light stage processing to enable single image portrait relighting~\cite{pandey2021total, sun2019single, zhou2019deep, wang2020single, weir2022deep, nestmeyer2020learning, kim2024switchlight, legendre2020learning}. Novel view synthesis and animation along with relighting are present with the development of 3D representations such as mesh~\cite{lombardi2018deep, bi2021deep},  volumetric primitive~\cite{lombardi2021mixture, yang2023towards, yang2024vrmm}, NeRF~\cite{xu2023renerf, sarkar2023litnerf, rao2022vorf} and 3D Gaussian~\cite{saito2024relightable, he2024diffrelight}. Inverse rendering is widely used to decouple intrinsics and enable lighting manipulation~\cite{nestmeyer2020learning, kim2024switchlight, pandey2021total, tan2022volux, zhang2024learning}. Recent development of diffusion models enables more lighting condition interfaces like prompt~\cite{iclight}, background~\cite{ren2024relightful}, and scribble~\cite{mei2023lightpainter}.

While the expensive studio-captured data provides ground truth supervision, easily accessible casual image collections offer more diversity and scalability. Deep generative neural networks like GANs are usually adopted to learn the priors of appearance and lighting. Early works explore style transfer~\cite{abdal2021styleflow, shih2014style, shu2017portrait, deng2020disentangled, fu2024hierarchical} to transfer lighting, but fail to maintain identity consistency. The quotient image~\cite{shashua2001quotient, peers2007post} decouples albedo from appearance and is widely used for plausible relighting~\cite{tewari2020stylerig, pan2021shading, jiang2023nerffacelighting}. Inverse rendering is also explored~\cite{deng2024lumigan} with self-supervised regulation. Relighting with novel view synthesis from casual images~\cite{sun2021nelf, tewari2020stylerig, zhang2021neural} is also present with semantic control over GANs. The incorporation of generative NeRF allows for 3D-aware portrait relighting~\cite{pan2021shading, tan2022volux, deng2024lumigan, mei2024holo, rao2024lite2relight, cai2024real, ranjan2023facelit, cai2024single}. 

\section{Method}
In the following, we first introduce preliminaries including spherical harmonics, precomputed radiance transfer, and 3D Gaussian head synthesis. Then we discuss design choices in generative 3D-GAN relighting. Finally, we introduce the proposed Unified Neural Radiance Transfer and the loss functions.

\subsection{Preliminaries}
\paragraph{Precomputed Radiance Transfer} Computing global illumination is costly because it requires iterative path tracing due to multi-bounce reflections, refractions, and scattering of light. \citet{sloan2002precomputed} propose to precompute radiance transfer functions, which are applied to actual incident lighting at run-time. Assuming distant illumination $L_{\text{in}}$,  exit radiance $L_{\text{out}, p}$ from a point $p$ is computed by the rendering equation with light transport:
\begin{align}\label{equation::prt}
L_{\text{out}, p}(\hat{\mathbf{v}}) 
&= \int_{\Omega} L_{\text{in}}(\hat{\mathbf{l}}) V_p(\hat{\mathbf{l}}) f_r(\hat{\mathbf{v}}, \hat{\mathbf{l}}) \max(0, \hat{\mathbf{n}} \cdot \hat{\mathbf{l}}) \, d\hat{\mathbf{l}} 
\end{align}
where $\hat{\mathbf{n}}$ is the surface normal at $p$, $\hat{\mathbf{l}}$ and $\hat{\mathbf{v}}$ are global light direction and viewing direction, $f_r$ is Bidirectional Reflectance Distribution Function (BRDF) and $V_p$ is the light transport function considering self-occlusion and interreflection.

To efficiently compute the integral in Equation~\ref{equation::prt}, lighting and transfer functions are represented using $n$th-order spherical harmonics (SH). For specular surfaces, a transfer matrix $\mathbf{M}_p$ including the BRDF and cosine terms describes the radiance transfer via
\begin{equation}
    \mathbf{L}_{\text{out}, p} = \sum\nolimits_{j=1}^{n^2} \mathbf{M}_p^{ij} \mathbf{L}_{\text{in}}^j.
\end{equation}
Specifically for diffuse surfaces, radiance transfer can be represented by a transfer vector $\mathbf{V}_p$ producing scalar exit radiance $L_{\text{out}, p}$ by
\begin{equation}
    L_{\text{out}, p} = \sum\nolimits_{i=1}^{n^2} \mathbf{V}_p^i \mathbf{L}_p^i.
\end{equation}

In modern neural rendering, precomputing transfer functions is often impractical because the geometric and material properties are unknown. Previous works~\cite{saito2024relightable, deng2024lumigan, zhang2024prtgaussian, xing2024real} choose to parameterize and learn diffuse radiance transfer vectors, while employing alternative representations for specular terms because the transfer matrix is too large. We instead propose to use a unified representation that models both diffuse terms and low-frequency reflections.

\paragraph{3D Gaussian Head Synthesis} 
3D Gaussian Splatting~\cite{kerbl20233d} is recently introduced as a point-based 3D representation that assigns to each point 5 attributes: position $\mu\in\mathbb{R}^3$, scale $s\in\mathbb{R}^3$, rotation quaternion $q\in\mathbb{R}^4$, opacity $\alpha\in\mathbb{R}$ and color $c\in\mathbb{R}^3$. A Gaussian ellipsoid is formulated as
\begin{equation}
G(x) = e^{-\frac{1}{2}(x-\mu)^{\mathrm{T}} \Sigma^{-1} (x-\mu)},
\quad\text{with}~~\Sigma = RS S^{\mathrm{T}} R^{\mathrm{T}},
\end{equation}
where $x$ is the world coordinates, and the scaling matrix $S$ and rotation matrix $R$ are derived from the scaling $s$ and quaternion $q$. The color $C$ of a pixel in image space is computed by blending $N$ overlapping sorted points:
\begin{equation}
C = \sum\nolimits_{i=1}^{N} c_i \alpha_i \prod\nolimits_{j=1}^{i-1} (1 - \alpha_j).
\end{equation}
Recent works in 3D Gaussian head synthesis~\cite{hyun2024adversarial, kirschstein2024gghead} allow for native high-resolution head synthesis in real time. \citet{kirschstein2024gghead} bind 3D Gaussians to a template mesh and use StyleGAN2~\cite{karras2020analyzing} to generate UV maps of each Gaussian attribute. \citet{hyun2024adversarial} instead explore a hierarchical generative paradigm, where the 3D Gaussians at each layer are constrained by anchor points from the previous layer.

\subsection{Unified Neural Radiance Transfer}
Radiance transfer was primarily used to describe the diffuse component of lighting. We show that by incorporating view directions into the generative framework, a Unified Neural Radiance Transfer representation is capable of describing both diffuse and low-frequency specular terms.
Revisiting Equation~\ref{equation::prt}, by absorbing the light transport, BRDF, and cosine term into a unified radiance transfer function $T_p$ in SH basis,
the rendering equation becomes
\begin{equation}
\label{equation::unified_radiance_transfer}
L_{\text{out}, p}(\hat{\mathbf{v}}) 
=\int_{\Omega} L_{\text{in}}(\hat{\mathbf{l}}) T_p(\hat{\mathbf{v}}, \hat{\mathbf{l}}) \, d\hat{\mathbf{l}} \approx\sum\nolimits_{i=1}^{n^2} l_{\text{in}}^{i} t^{i}_p(\hat{\mathbf{v}}),
\end{equation}
where $l^{i}_{\text{in}}$ and $t^{i}_p(\hat{\mathbf{v}})$ are the $n$th-order expansion coefficients. Specifically, we disentangle the albedo $\rho_p$ from $t^{i}_p(\hat{\mathbf{v}})$ for disentangling appearance from lighting. 
With minor notation abuse, the view-dependent exit radiance is formulated as
\begin{equation}
\label{equation::light_transport}
    L_{\text{out}, p}(\hat{\mathbf{v}}) = \rho_p\sum\nolimits_{i=1}^{n^2} l_{\text{in}}^{i} t^{i}_p(\hat{\mathbf{v}}).
\end{equation}
In this way, both diffuse and specular terms can be handled in a unified manner by making radiance transfer coefficients a function of view direction. 

Based on 3D Gaussian GAN framework, we additionally assign each Gaussian point $p$ an albedo $\mathbf{\rho}_p\in [0,1]^3$ and a transfer vector $\mathbf{t}_p\in\mathbb{R}^{3\times n^2}$ to replace the original view-independent color. Specifically, as shown in Figure~\ref{fig::pipeline}, the generator gives each Gaussian point $p$ an embedding $\mathbf{x}_p$. The transfer vector $\mathbf{t}_p$ 
is obtained by concatenating the Gaussian embedding with encoded view direction and sending it to a transfer vector decoder $\Psi_t$,
\begin{equation}
    \mathbf{t}_p(\hat{\mathbf{v}}) = \Psi_{t}(\mathbf{x}'_p || \mathbf{v}_p), 
\end{equation}
where $\mathbf{x}'_p$ is a shortened embedding transformed linearly from the Gaussian embedding $\mathbf{x}_p$. It can be stored to perform real-time rendering with standard 3D Gaussian splatting after generation.

Without explicit supervision of albedo, the ambiguity between albedo and transfer vectors is unavoidable, as demonstrated in Section~\ref{sec::ablation}. 
Observing that the illumination in most real images is approximately white light, we propose to train the model under white light conditions. The light and transfer vectors are then reduced to single channel, i.e. $\mathbf{l}, \mathbf{t}\in\mathbb{R}^{1\times n^2}$. In this way, the contribution of the transfer vector to the color distribution is discarded, and thus the ambiguity is resolved. We find in the experiment that adversarial training will average out the impact of light color residuals on the albedo. Moreover, due to the linearity of our unified radiance transfer functions, our model can render colored lighting at test-time by channel-wise applying Equation~\ref{equation::light_transport}.

\section{Experiments}

\subsection{Implementation Details}
Our method is based on GSGAN~\cite{hyun2024adversarial}, which follows the same data preparation pipeline as EG3D~\cite{chan2022efficient}. 
The lighting labels, represented as monochromatic 3rd-order spherical harmonics, are extracted using DPR~\cite{zhou2019deep}, following the approach of~\citet{jiang2023nerffacelighting}.
We use MODNet~\cite{ke2022modnet} to extract foreground components to alleviate background leakage. The transfer vector decoder $\Psi_t$ is implemented as a lightweight MLP with one hidden layer of 64 units. We train the models on the FFHQ dataset~\cite{karras2019style} from scratch for 15M images with four NVIDIA H800 GPUs.

\subsection{Quantitative Comparisons}
\label{sec::quantitative}
\paragraph{Baselines.} We compare our method with previous publicly accessible works for novel view portrait relighting: ShadeGAN~\cite{pan2021shading}, Volux-GAN~\cite{tan2022volux}, NeRFFaceLighting~\cite{jiang2023nerffacelighting}. We also include LumiGAN~\cite{deng2024lumigan}, which is not accessible, and report results from the original paper if presented.

\paragraph{Generative quality and rendering speed.}  We follow ~\citet{chan2022efficient} and compute Frechnet Inception Distance (FID)~\cite{heusel2017gans} and Kernel Inception Distance (KID)~\cite{binkowski2018demystifying} between 50k generated images and all training images. From Table~\ref{tab::generative_quality}, we see that EG3D-based methods have a higher FID due to the 2D super-resolution module, which upsamples low-resolution volume rendering results at the cost of losing 3D consistency. Our method achieves similar or even better generative quality than our backbone GSGAN~\cite{hyun2024adversarial}. Moreover, our method maintains a high rendering speed, surpassing other relighting works by at least 12$\times$. This is attributed to the simple yet effective architecture of our unified radiance transfer relighting framework.

\begin{table}[t]
\caption{Comparison of generative quality and rendering speed.  KID is reported by 100$\times$. SR denotes super resolution. For relighting methods, rendering speed is measured under varying illumination and view direction. All methods render at 512 resolution except for Volux-GAN at 256 on a single RTX A6000 GPU. We bold the best score, underline the second, and double-underline the third.}
\label{tab::generative_quality}
\centering
\resizebox{\linewidth}{!}{
    \begin{tabular}{lcccc}
    \toprule
    Method & SR  & FID $\downarrow$ & KID $\downarrow$  & Rendering Speed $\uparrow$ \\
    \midrule
    EG3D & \checkmark  & \underline{4.30} & \textbf{0.132} &  \underline{\underline{42}} fps \\
    GSGAN & $\times$  & 5.60 & 0.161 & \textbf{245} fps \\
    
    \midrule
    Volux-GAN & \checkmark & 59.79 & 4.124  & 21 fps \\
    ShadeGAN & \checkmark & 9.92 & 0.453 & 0.5 fps\\
    LumiGAN & \checkmark & \underline{\underline{5.28}} & 0.251  & - \\
    NeRFFaceLighting & \checkmark & \textbf{4.16 } & \underline{0.147}  & 2.8 fps \\
    \midrule
    Ours & $\times$ & 5.71 & \underline{\underline{0.151}}  & \underline{243} fps\\
    \bottomrule
    \end{tabular}
}
\end{table}

\begin{table}[t]
    \caption{Comparison of lighting quality.}
    \label{tab::lighting}
    \centering
    \begin{tabular}{lcc}
        \toprule    
        Method  & Lighting Error$\downarrow$ & Lighting Stability$\downarrow$  \\
        \midrule
        StyleFlow & 0.7523 & 0.1530 \\
        DisCoFaceGAN & \textbf{0.5860} & \textbf{0.1335} \\
        GAN-Control & 0.6647 & 0.1485  \\
        \midrule
        ShadeGAN & 1.0714 & 0.2149 \\
        NeRFFaceLighting & 0.6377 & \underline{0.1455}  \\
        \midrule
        Ours & \underline{0.6213} & 0.1537 \\
        \bottomrule
    \end{tabular}
\end{table}

\begin{table}[t]
    \caption{Comparison of spatial consistency. SR denotes super resolution.}
    \label{tab::spatial}
    \centering
    \resizebox{\linewidth}{!}{
    \begin{tabular}{lcccc}
        \toprule
        
        Method  & SR & PSNR$\uparrow$ & SSIM$\uparrow$ & Id. Sim. (view) $\uparrow$ \\
        \midrule
        EG3D & \checkmark & \textbf{38.13} & \textbf{0.9612}  & 0.7501  \\
        GSGAN & $\times$ &35.85 & 0.9395  &  0.7416 \\
        \midrule
        ShadeGAN & \checkmark & 24.23 & 0.7316 & 0.5238 \\
        Volux-GAN & \checkmark & 31.64 & 0.8820 & 0.6882 \\
        NeRFFaceLighting & \checkmark & 37.98 & 0.9561 & 0.7303 \\
        \midrule
        Ours & $\times$ & 35.57 & 0.9521 & \textbf{0.7538} \\
        \bottomrule
    \end{tabular}
    }
\end{table}

\paragraph{Lighting quality.}  Following~\citet{jiang2023nerffacelighting}, we measure the relighting performance with an off-the-shelf estimator~\cite{feng2021learning} using two metrics: the \emph{Lighting Error} computes the distance between real lighting and estimated lighting in fake images; the \emph{Lighting Stability} computes the standard deviation of estimated lighting in 100 fake images generated using the same lighting conditions. The metrics are averaged for 1000 samples. Since these metrics can only be applied to methods with lighting conditions estimated from real images, Volux-GAN and LumiGAN are excluded. Instead, StyleFlow~\cite{abdal2021styleflow}, DisCoFaceGAN~\cite{deng2020disentangled}, and GAN-Control~\cite{shoshan2021gan} are included as 2D generative relighting methods for comparison. Table~\ref{tab::lighting} shows that our method is comparable in lighting quality with state-of-the-art 3D-portrait relighting methods. The 2D relighting method DisCoFaceGAN has higher lighting scores at the cost of low generative quality. Moreover, our method can perform RGB relighting, thanks to its linear radiance transfer framework. This is not achievable by other methods based on implicit neural relighting.

\paragraph{Spatial consistency.} 
Following~\citet{hyun2024adversarial}, we use a surface estimation model NeUS2~\cite{wang2023neus2} to validate the spatial consistency of 3D-aware relighting models. Specifically, we randomly generate 30 views of a lit subject, fit them to surface estimation, and compute the reconstruction error using standard PSNR and SSIM metrics. We generate 30 subjects and record the average metrics. We also include the identity similarity metric proposed by~\citet{tan2022volux}, and report the mean score across 30 views. From Table~\ref{tab::spatial} we see that although EG3D-based methods do not preserve exact 3D consistency due to 2D super resolution, they do not exhibit disadvantages in terms of reconstruction error. This is possibly because EG3D and NeUS2 share a similar implicit representation and volume rendering process. However, in terms of identity similarity, our method achieves the highest score, demonstrating better 3D consistency than other methods. In general, EG3D-based methods suffer from texture flickering when continuously changing viewpoint due to 2D super resolution, while our method renders an explicit 3D Gaussian model and maintains view consistency. We further refer to the video demo for better visual comparison.

\begin{figure*}[t]
\renewcommand{\arraystretch}{0}%
\setlength\tabcolsep{0pt}
\setlength{\defaultaddspace}{0.3em}
\centering
    \begin{tabular}{cccc}
    \raisebox{2.2\normalbaselineskip}[0pt][0pt]{\rotatebox[origin=c]{90}{ShadeGAN}}&
    \includegraphics[width=0.33\linewidth]{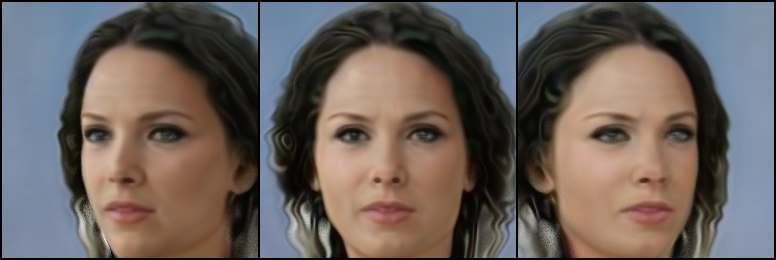} &
    \includegraphics[width=0.33\linewidth]{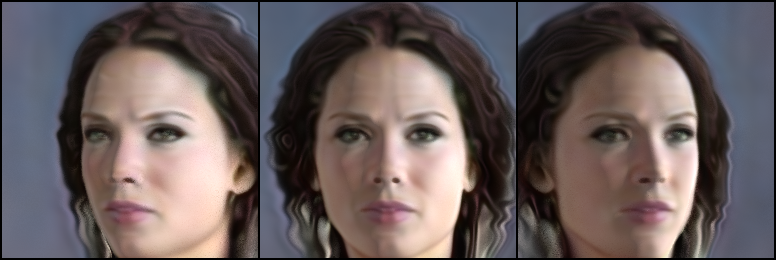} &
    \includegraphics[width=0.33\linewidth]{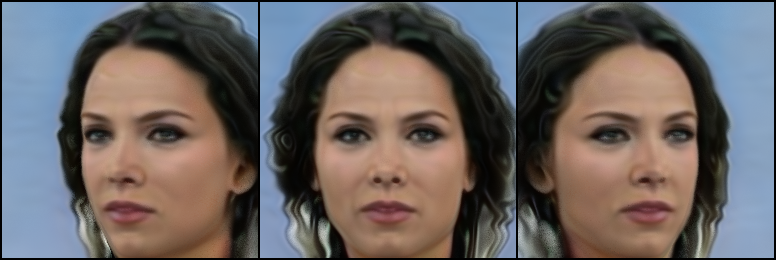} \\
    \addlinespace
    \raisebox{2.2\normalbaselineskip}[0pt][0pt]{\rotatebox[origin=c]{90}{Volux-GAN}}&
    \includegraphics[width=0.33\linewidth]{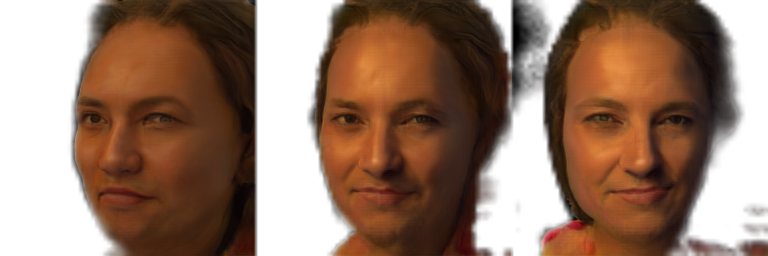} &
    \includegraphics[width=0.33\linewidth]{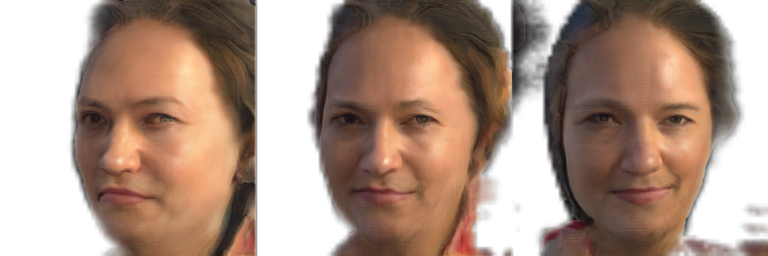} &
    \includegraphics[width=0.33\linewidth]{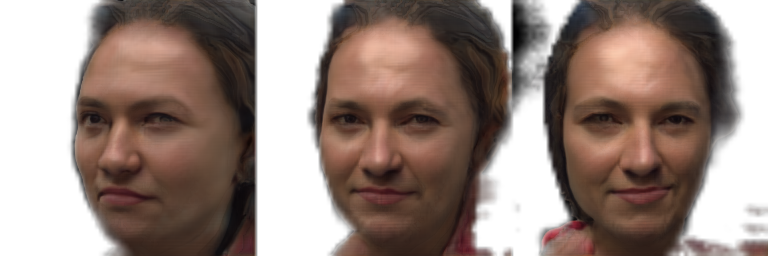} \\
    \addlinespace
    \raisebox{2.3\normalbaselineskip}[0pt][0pt]{\rotatebox[origin=c]{90}{NeRFFaceLighting}}&
    \includegraphics[width=0.33\linewidth]{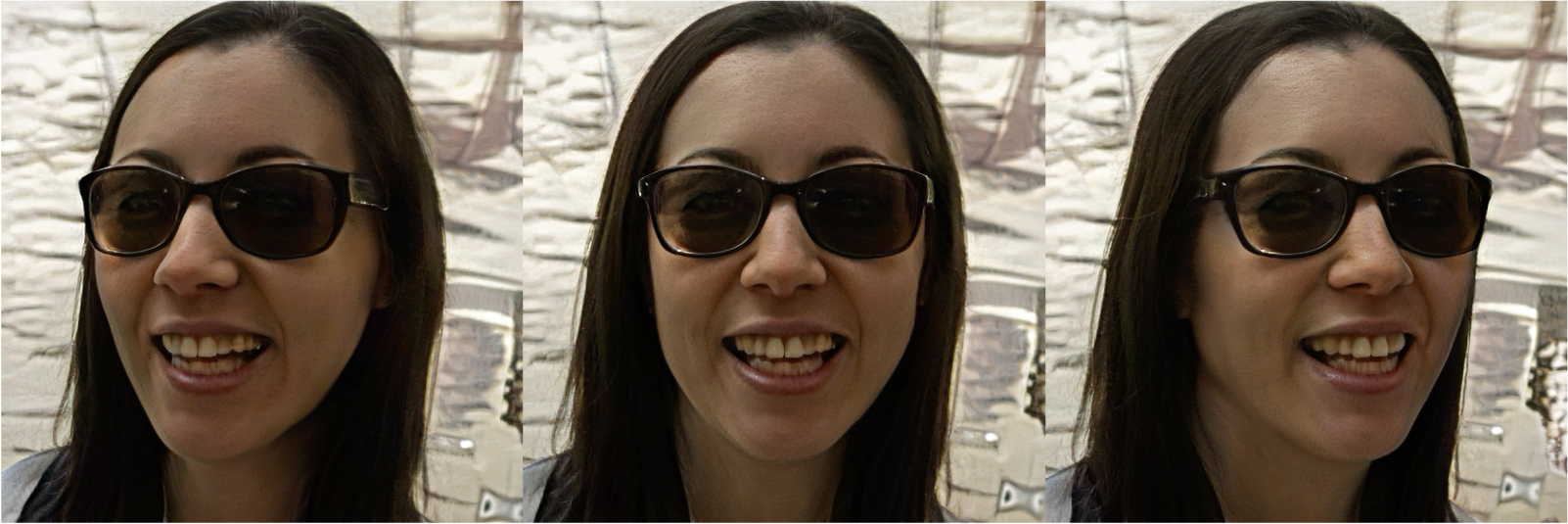} &
    \includegraphics[width=0.33\linewidth]{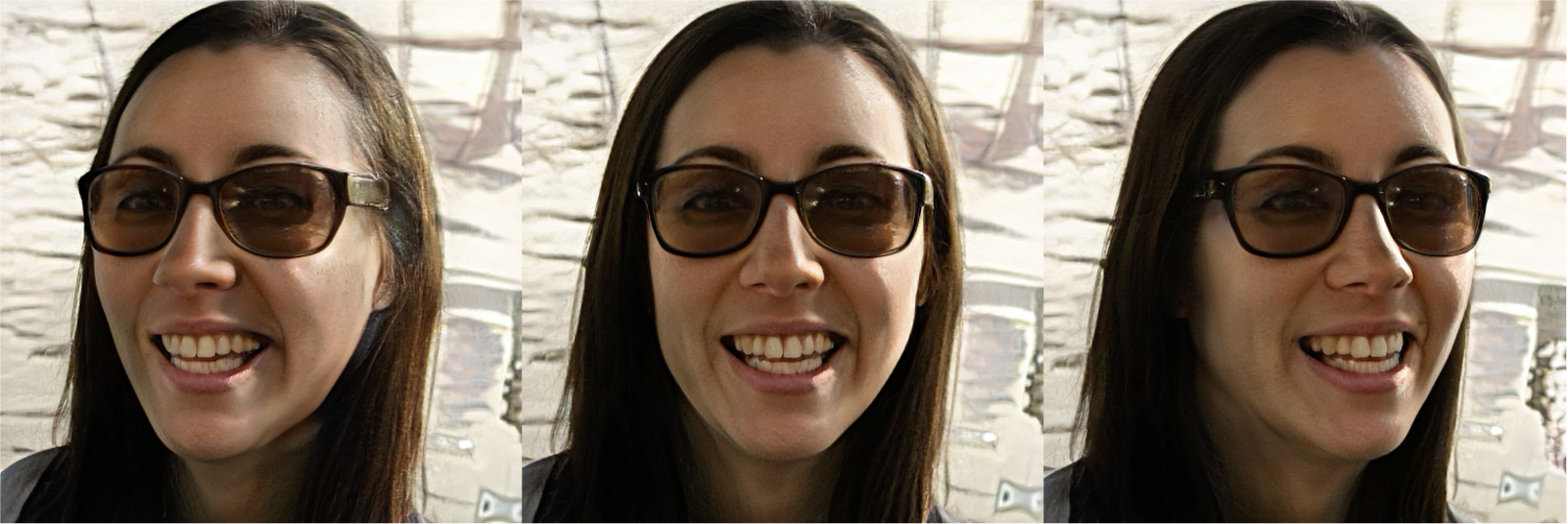} &
    \includegraphics[width=0.33\linewidth]{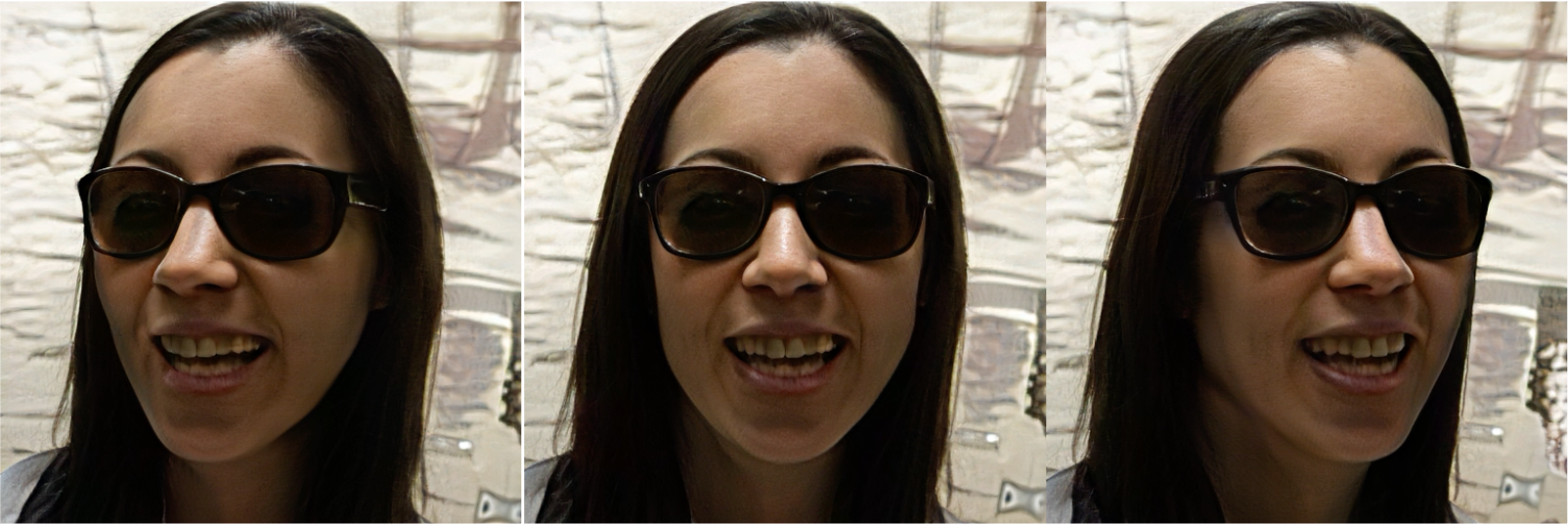} \\
    \addlinespace
    \raisebox{2.2\normalbaselineskip}[0pt][0pt]{\rotatebox[origin=c]{90}{LumiGAN}}&
    \includegraphics[width=0.33\linewidth]{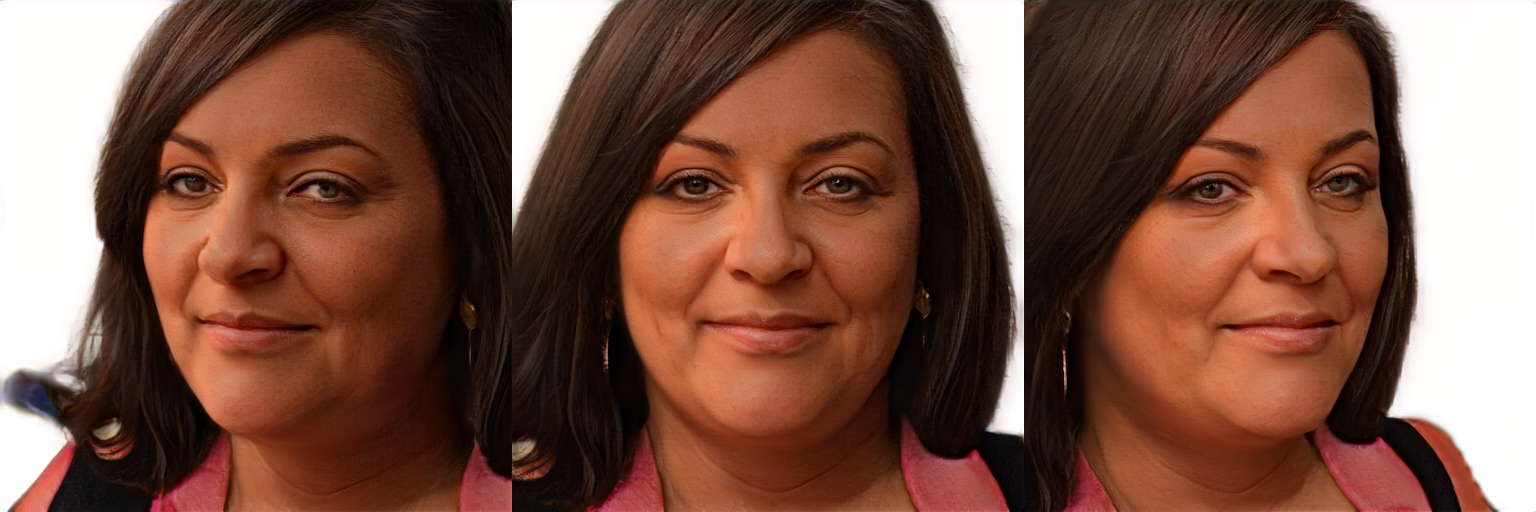} &
    \includegraphics[width=0.33\linewidth]{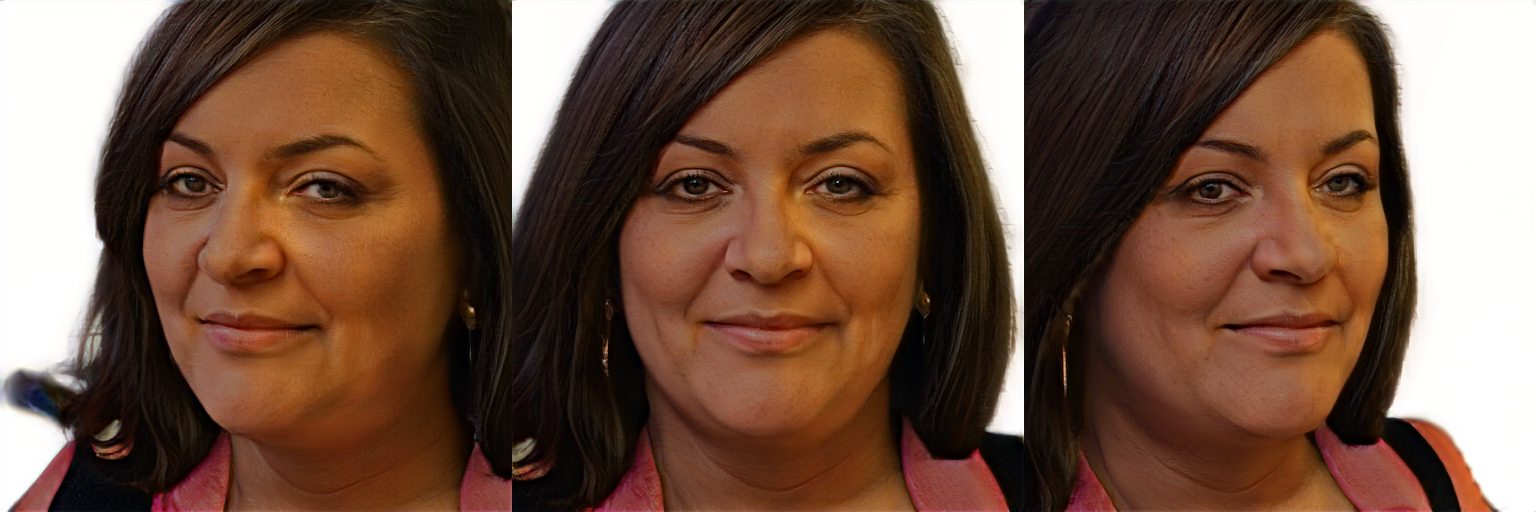} &
    \includegraphics[width=0.33\linewidth]{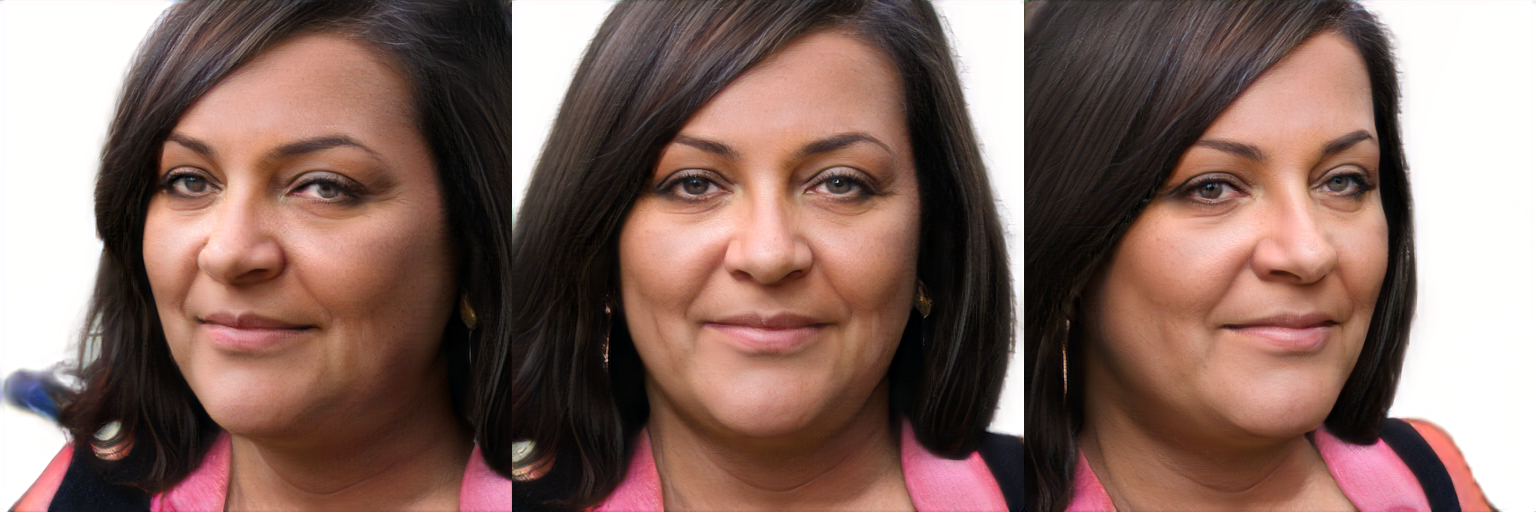} \\
    \addlinespace
    \raisebox{2.2\normalbaselineskip}[0pt][0pt]{\rotatebox[origin=c]{90}{NeLF}}&
    \includegraphics[width=0.33\linewidth]{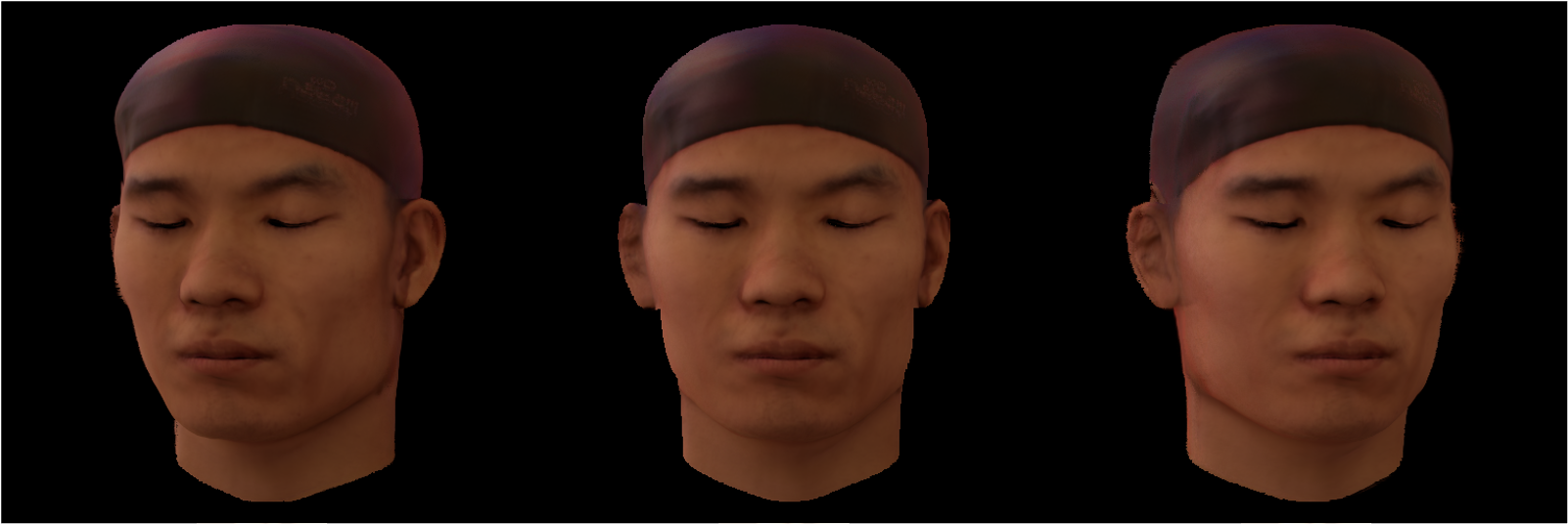} &
    \includegraphics[width=0.33\linewidth]{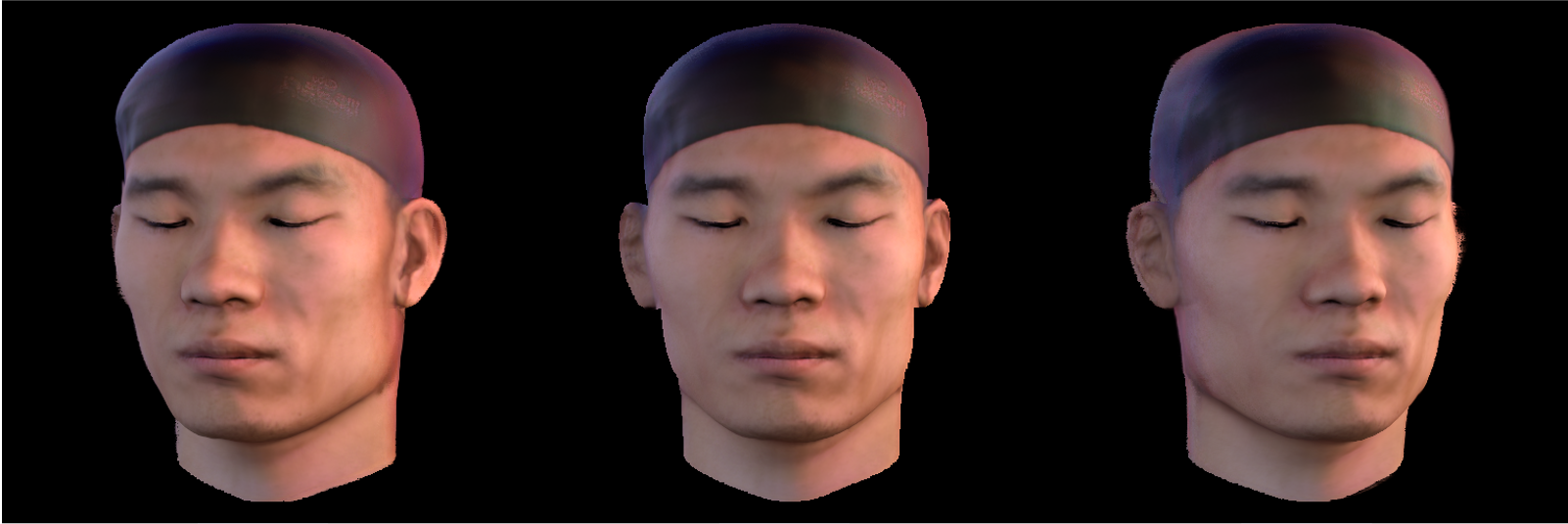} &
    \includegraphics[width=0.33\linewidth]{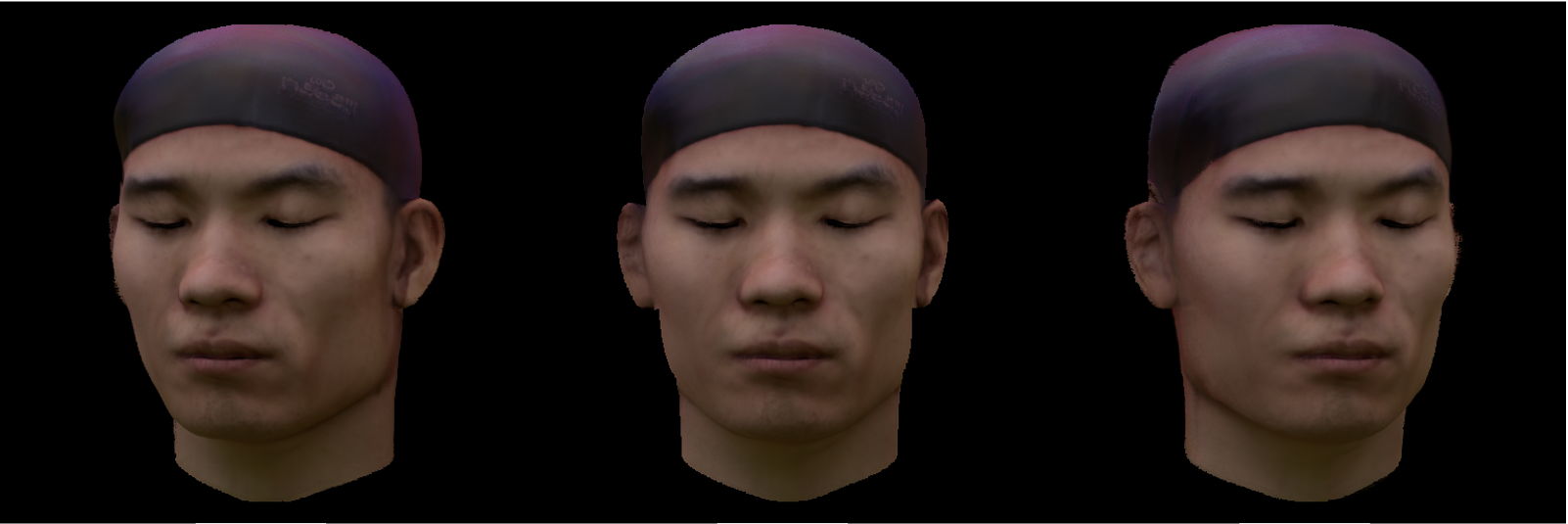} \\
    \addlinespace
    \raisebox{2.2\normalbaselineskip}[0pt][0pt]{\rotatebox[origin=c]{90}{Ours}}&
    \includegraphics[width=0.33\linewidth]{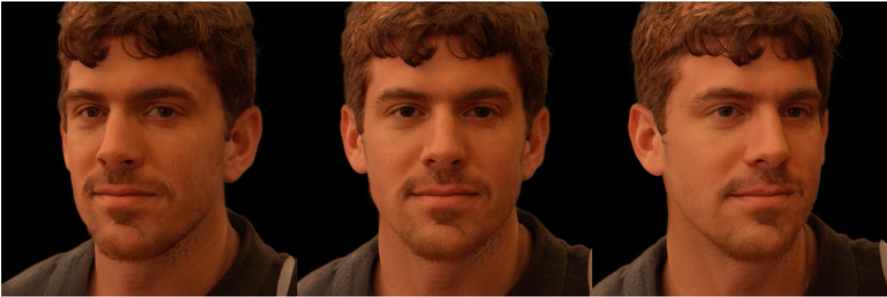} &
    \includegraphics[width=0.33\linewidth]{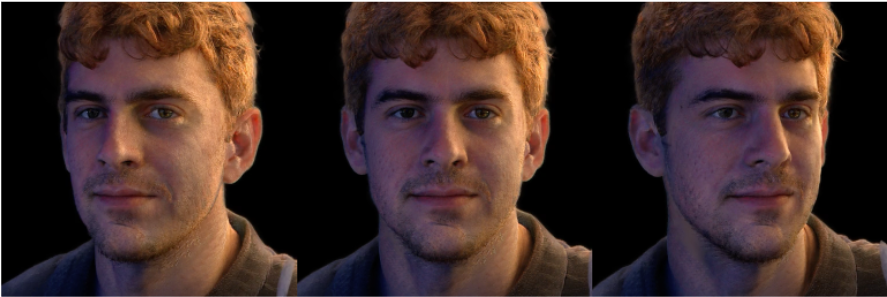} &
    \includegraphics[width=0.33\linewidth]{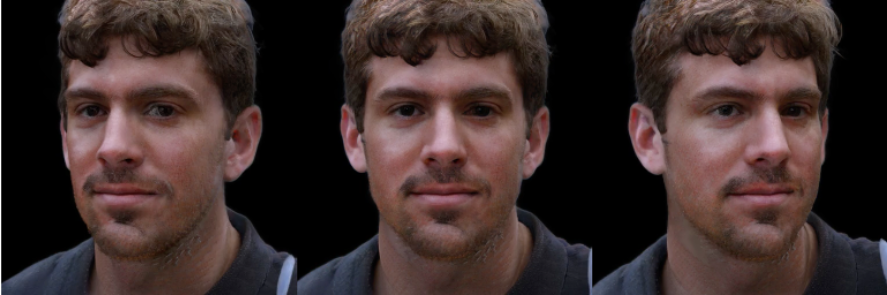} \\
    \addlinespace
    \raisebox{2.2\normalbaselineskip}[0pt][0pt]{\rotatebox[origin=c]{90}{Lighting}}
    &
    \includegraphics[width=0.33\linewidth,height=0.11\linewidth]{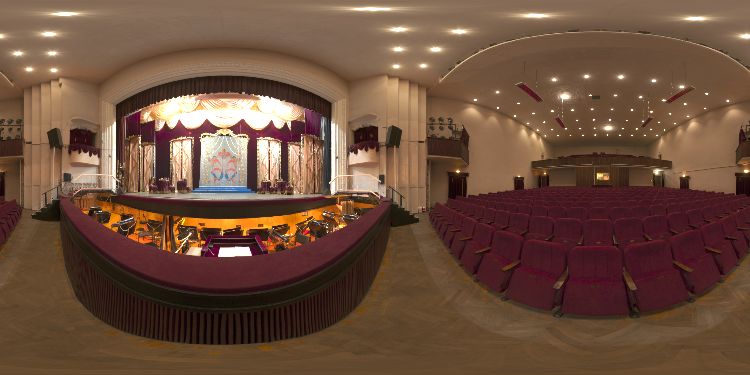} &
    \includegraphics[width=0.33\linewidth,height=0.11\linewidth]{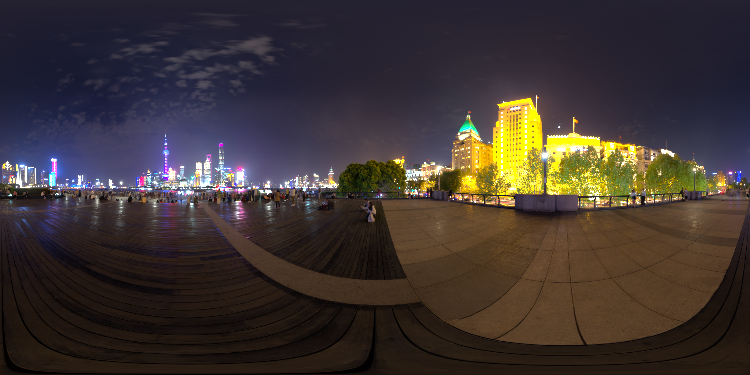} &
    \includegraphics[width=0.33\linewidth,height=0.11\linewidth]{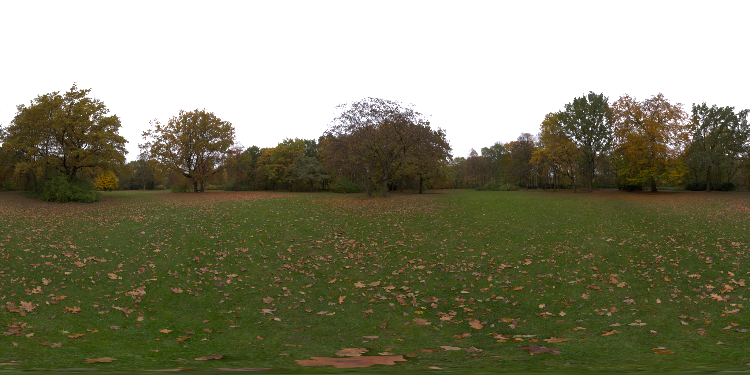}\\
    \end{tabular}
\caption{\textbf{Qualitative Comparison. } We use the same environment maps as LumiGAN~\cite{deng2024lumigan} for comparison, since it is not publicly accessible.
}
\label{fig::envmap_compare}
\end{figure*}

\subsection{Qualitative Comparison}
The Lighting Error and Lighting Stability metrics used in Section~\ref{sec::quantitative} can only evaluate the lighting similarity by an off-the-shelf lighting estimator. To further evaluate lighting quality under colored light conditions and compare the lighting realism under human eyes, we compare our method to previous works in real environment map lighting conditions. We include NeLF~\cite{sun2021nelf}, which is trained using a synthetic lightstage dataset and needs multi-view inputs. Figure~\ref{fig::envmap_compare} shows that our method achieves better relighting quality than previous methods. Volux-GAN~\cite{tan2022volux} uses 2D convolution layers to add shadows and details, leading to view inconsistencies. NeRFFaceLighting~\cite{jiang2023nerffacelighting} can only handle white-light conditions and performs poorly on environment map lighting. LumiGAN~\cite{deng2024lumigan} relies on inverse rendering, and incorrect geometry decoupling leads to lighting artifacts, like the unnatural lit parts on the cheek and beside the wings of the nose. NeLF shows poor visual quality and loses facial details like wrinkles and pores. In contrast, our method performs well both in visual quality and relighting quality.

\begin{figure*}[t]
    \centering        \includegraphics[width=0.9\linewidth]{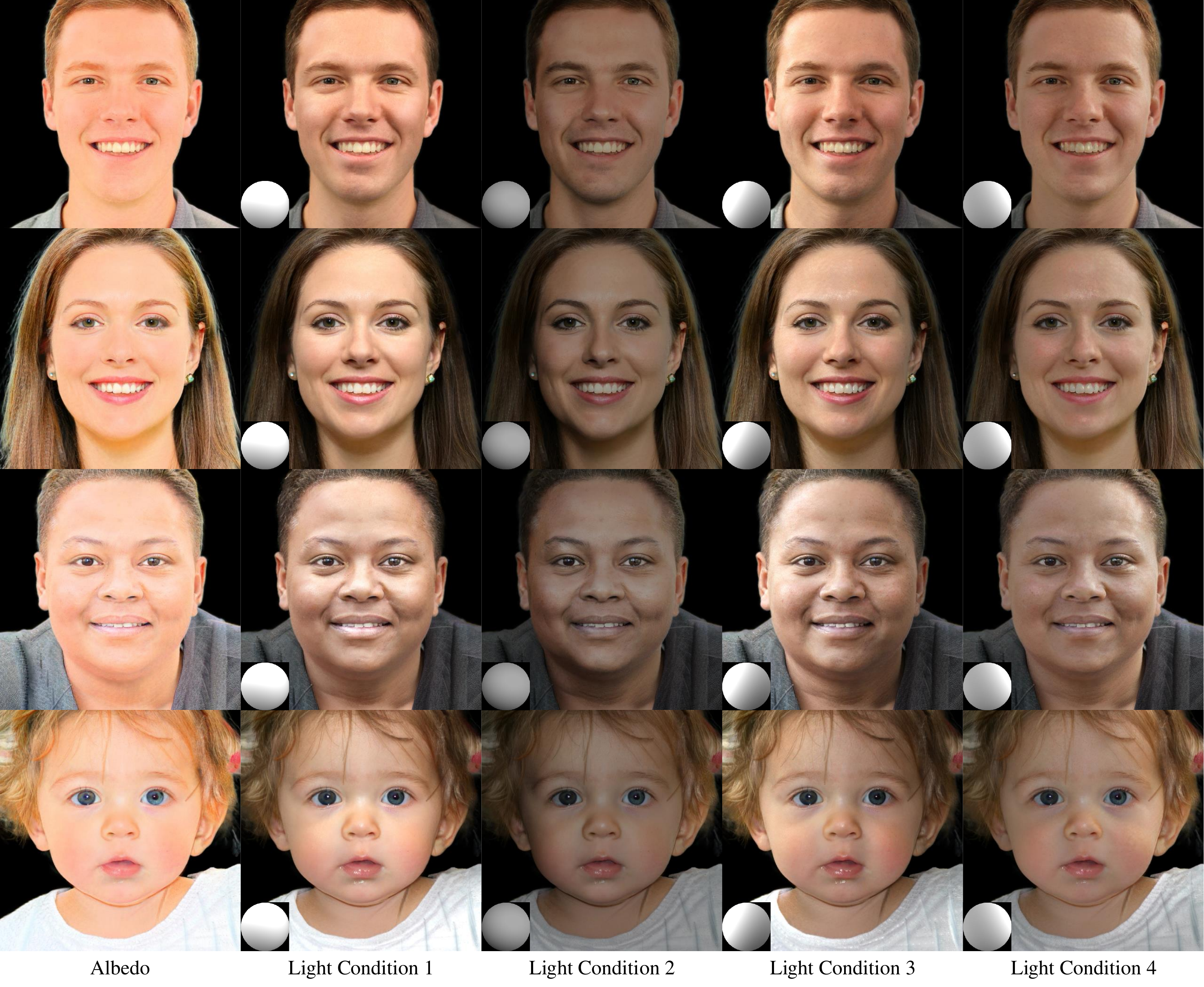}
    \caption{White lighting samples of our method. Lighting conditions are provided at the lower left corner.}
    \label{fig::white_light}
\end{figure*}

\subsection{Visual Relighting Results}
\begin{figure}[t!]
    \centering        \includegraphics[width=1.0\linewidth]{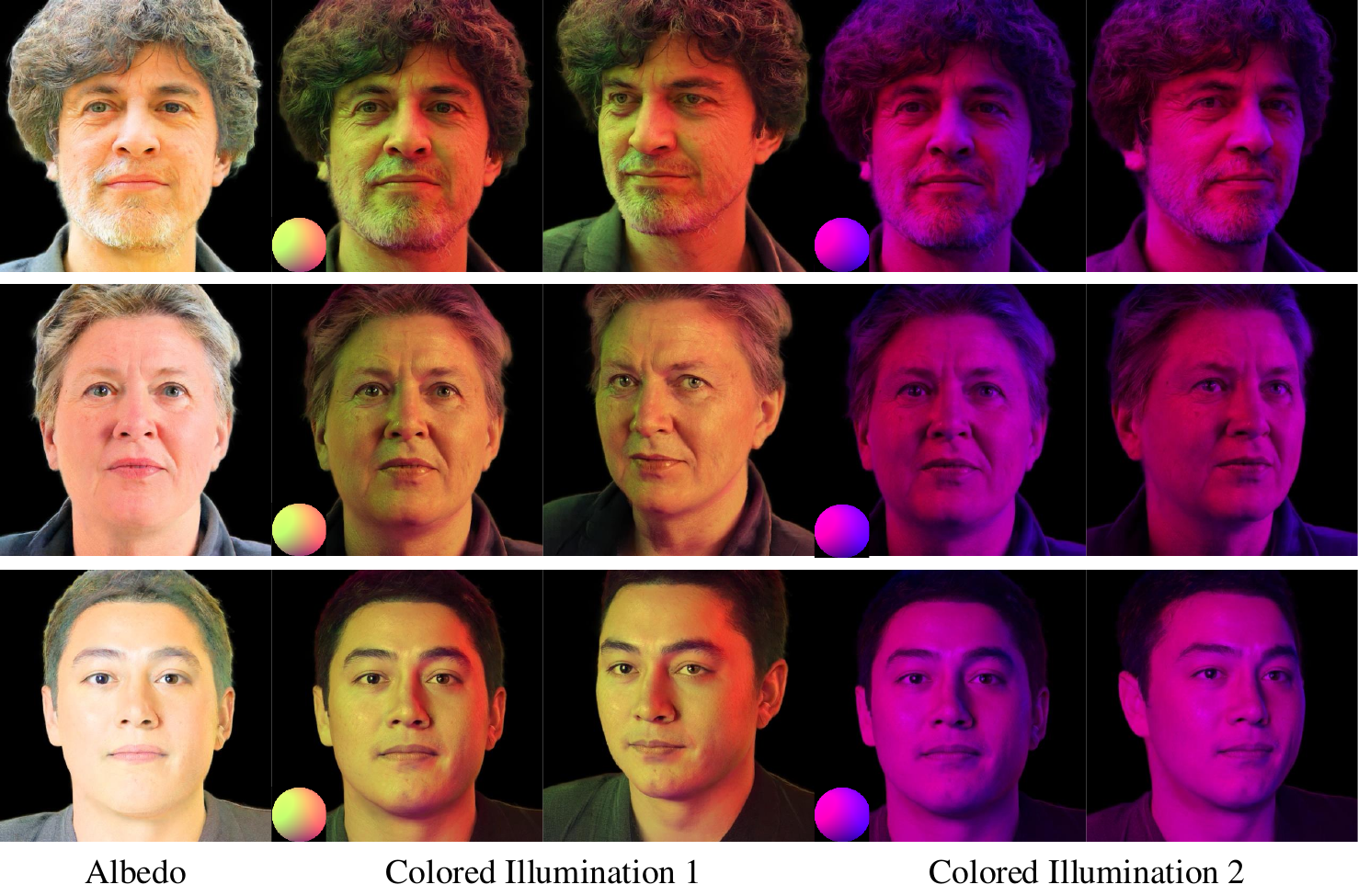}
    \caption{RGB lighting samples of our method.}
    \label{fig::rgb_light}
\end{figure}

\begin{figure*}[t]
    \centering    \includegraphics[width=0.9\linewidth]{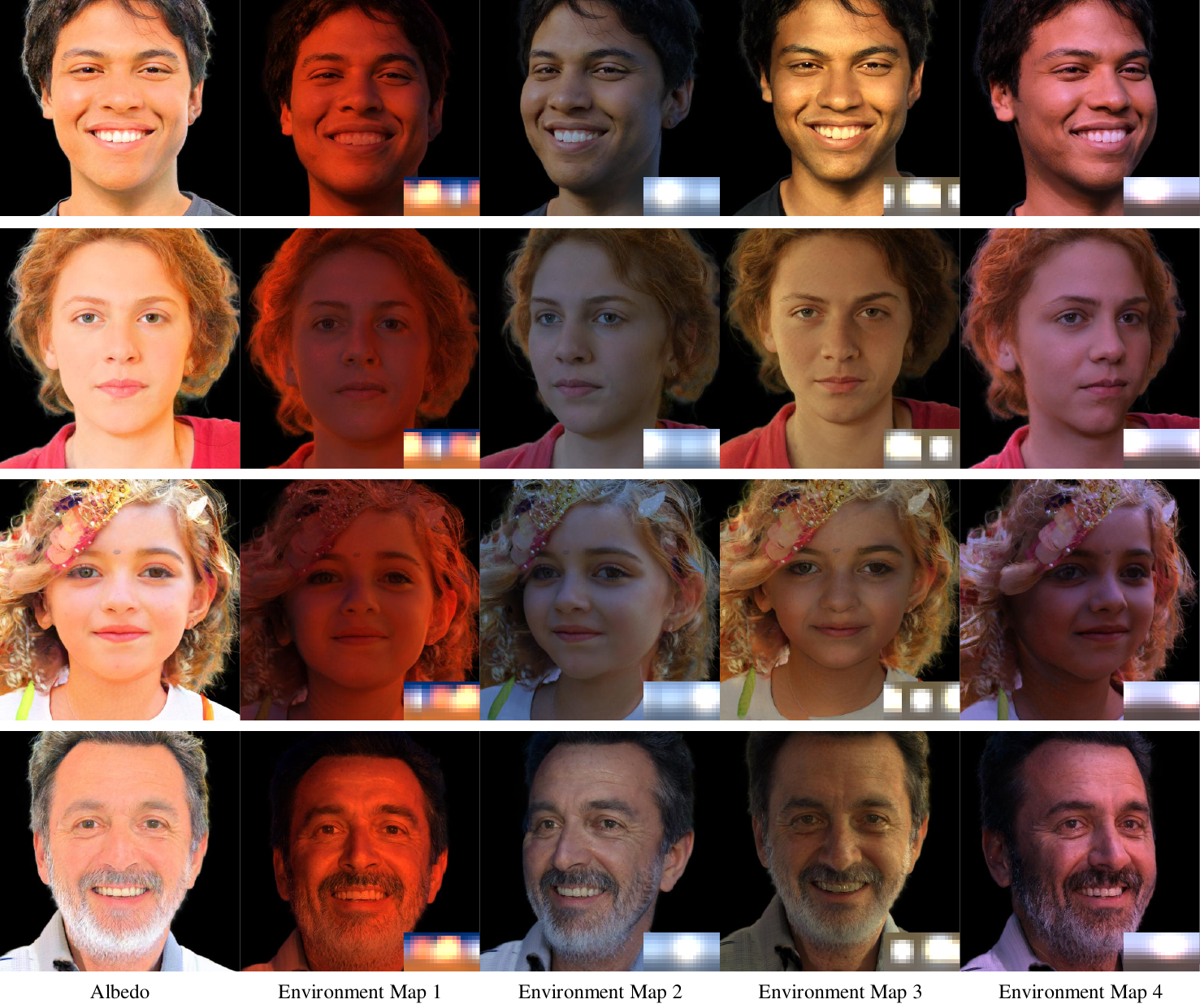}
    \caption{Environment map relighting samples of our method. The first column presents albedo images and the rest columns present renderings under different environment maps and view directions. Environment maps are rotated for different rows.}
    \label{fig::env_map}
\end{figure*}

Our model presents overall good photorealism under different light intensity and directions, as shown in Fig~\ref{fig::white_light}. Colored illumination can also be handled correctly, as shown in Figure~\ref{fig::rgb_light}. Given an environment map, image-based lighting requires first converting it to spherical harmonic (SH) coefficients. Specifically, we apply Gaussian blurring before SH sampling with a radius of 0.1$\times$ image width to remove high-frequency details. Image-based lighting results are shown in Figure~\ref{fig::env_map}. Thanks to the unified radiance transfer framework, our approach responds linearly to incident light and innately satisfies the light transport consistency. Our method maintains 3D consistency in head poses while delivering plausible albedo and relighting results under various environment maps. 

\begin{figure*}[t]
    \centering        \includegraphics[width=\linewidth]{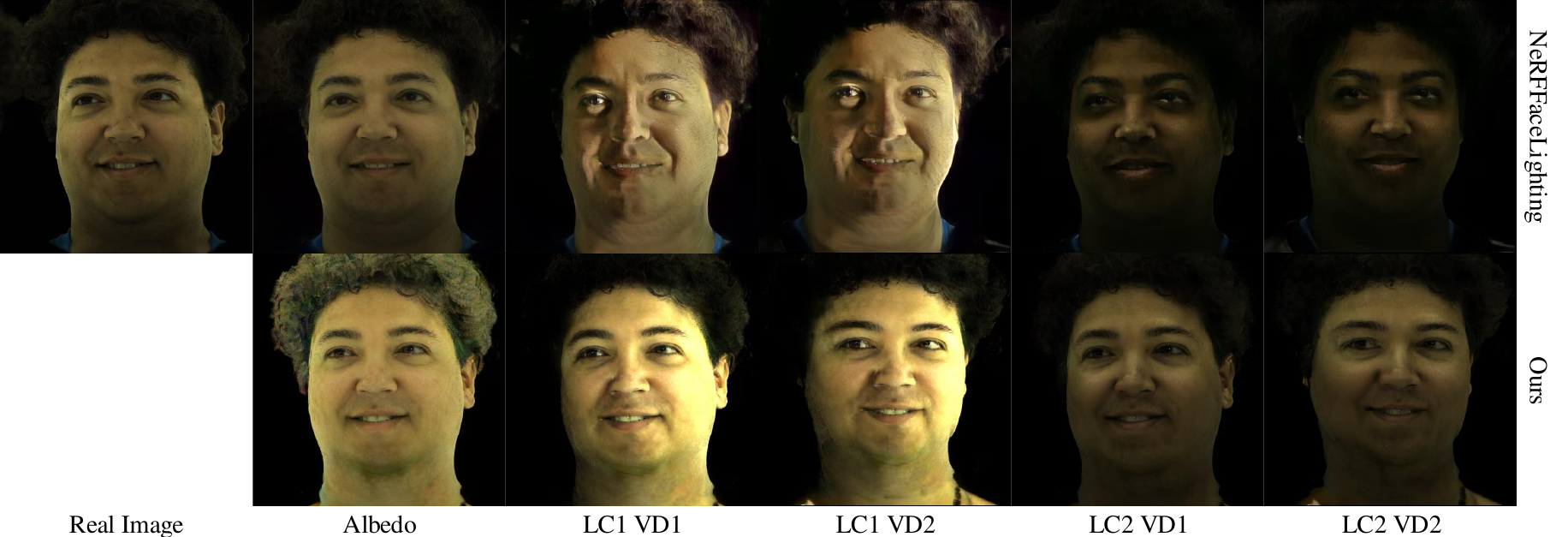}
    \caption{Results of real portrait relighting and novel view synthesis. LC and VD stand for light condition and view direction, respectively. Our method shows better photorealism and identity consistency compared to NeRFFaceLighting~\cite{jiang2023nerffacelighting}.}
    \label{fig::inversion}
\end{figure*}

\paragraph{Real portrait relighting}
With the help of the GAN inversion technique, our model can relight real-face images. The basic idea is to project a real-face image into the latent space and manipulate lighting during reconstruction. We implement the inversion similar to Pivotal Tuning~\cite{roich2022pivotal} and compare the result with NeRFFaceLighting in Figure~\ref{fig::inversion}. Our method produces better photorealism and identity consistency. 
We also quantitatively evaluate identity consistency using cosine similarity of VGGFace~\cite{parkhi2015deep} embedding on a subset of the Goliath dataset~\cite{martinez2024codec}, which has 4 sets of light-stage head capture. Our method (0.443) significantly outperforms NeRFFaceLighting (0.298).

\subsection{Other Design Choices and Ablation Study}\label{sec::ablation}

Multiple design choices exist for implementing unified radiance transfer in the 3D-GAN framework. First, the model could be trained using RGB lighting labels instead of monochromatic ones. Second, diffuse and specular reflectance could be modeled separately. We discuss these designs and validate their feasibility. We then conduct an ablation study on our proposed unified radiance transfer.

\begin{figure}[t]
    \centering       \includegraphics[width=0.75\linewidth]{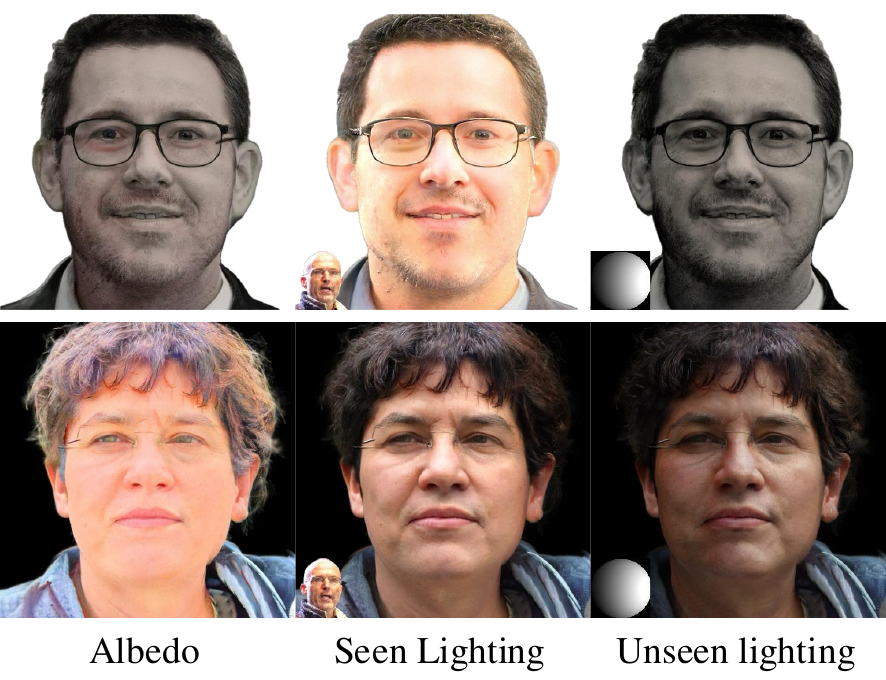}
    \caption{Visual comparison between training under RGB light condition (left) and white light condition (right), both trained for 3M images. We present albedo and relighting results under seen lighting and unseen lighting. Lighting conditions are provided at the lower left corner.}
    \label{fig::designs_1}
\end{figure}

\begin{figure}[t]
    \centering  \includegraphics[width=\linewidth]{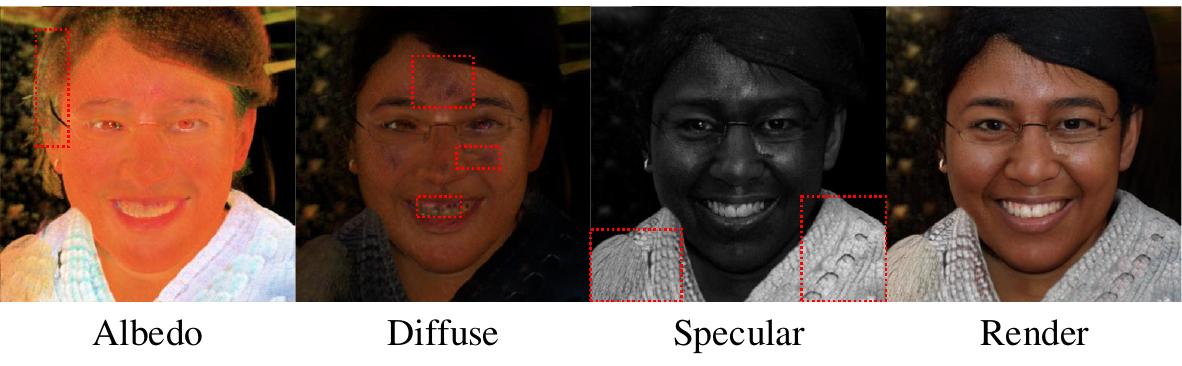}
    \caption{Visual demonstration of diffuse and specular decomposition. Color inconsistencies are marked by red dashed boxes.}
    \label{fig::designs_2}   
\end{figure}

\begin{figure}[t]
    \centering  \includegraphics[width=\linewidth]{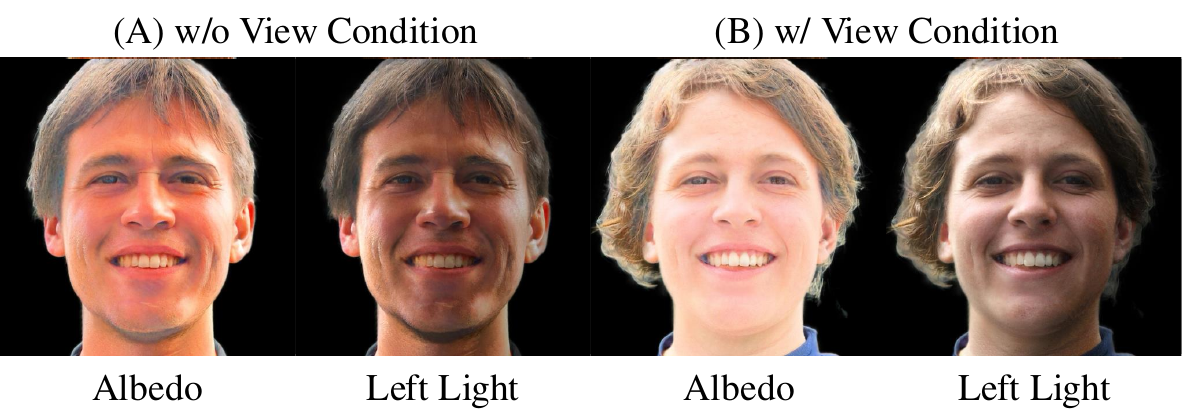}
    \caption{Visual ablation of view condition. Albedo and left light renderings with and without view condition are presented. Zoom in for a better view.}
    \label{fig::ablation}   
\end{figure}

\paragraph{No white-light assumption} Some works~\cite{fei2023split, phongthawee2024diffusionlight} allow extracting estimated RGB lighting conditions from a single image, which makes training on colored illumination possible. We follow the same setting and modify the parameterization of lighting labels and transfer functions from 1 channel to 3 channels. We use~\cite{fei2023split} to extract HDRI environment maps and turn them into 3rd-order RGB spherical harmonics coefficients. As shown in Figure~\ref{fig::designs_1},  without explicit supervision of intrinsics, training under RGB lighting conditions leads to faded albedo. In this case, the contribution to appearance is entirely provided by lighting, and relighting becomes a style transfer process. The model overfits on certain lighting conditions, does not learn a physically consistent lighting model, and fails under unseen lighting.
\paragraph{diffuse and specular decomposition} 
Inverse rendering is widely adopted in portrait relighting~\cite{saito2024relightable, kim2024switchlight, deng2024lumigan} to infer physical properties such as geometry, materials, and lighting. Diffuse and specular terms are often decoupled using different lighting models. We validate this setting in the 3D-GAN framework. We decouple the transfer vector $\mathbf{t}(\hat{\mathbf{v}})$ into view-independent diffuse transfer vector $\mathbf{t}_{\text{diffuse}}$ and view-dependent specular transfer vector $\mathbf{t}_{\text{specular}}(\hat{\mathbf{v}})$. However, without explicit supervision, albedo, diffuse, and specular terms are disentangled incorrectly and exhibit significant inconsistencies, as shown in Figure~\ref{fig::designs_2}.

\paragraph{Ablation study}
We validate the effectiveness of unified radiance transfer in Table~\ref{tab::ablation}. Without radiance transfer, we assume the color is directly predicted by a decoder that takes in lighting conditions. All models are trained for 3M images before evaluation. We find that radiance transfer provides strong lighting manipulation ability, demonstrated by the significant increase in Lighting Error in the case without radiance transfer. At the same time, without the view condition, the lighting accuracy also drops since the vanilla vector-based radiance transfer cannot model specular effects. In this case, the albedo is often observed with specular residuals. In contrast, our complete unified radiance transfer representation achieves the best lighting manipulation accuracy and yields smooth albedos.
From Figure~\ref{fig::ablation}, view conditioning reduces most highlight residuals in albedo and yields more satisfactory relighting results.

\begin{table}[t]
    \caption{Quantitative ablation of unified radiance transfer. R.T. denotes radiance transfer and V.C. denotes view condition.}
    \label{tab::ablation}
    \centering
    \begin{tabular}{lcccc}
        \toprule
        
        Method   & Lighting Error$\downarrow$ & Lighting Stability$\downarrow$  \\
        \midrule
        w/o R.T.   & 1.2296 & 0.4288 \\
        w/o V.C.  & 0.7019 & 0.1723 \\
        Ours   & \textbf{0.6520} & \textbf{0.1582} \\
        \bottomrule
    \end{tabular}
\end{table}

\section{Conclusion, Limitation, and Future Work}
\begin{sloppypar}
We present GSHeadRelight, a real-time relightable 3D-aware portrait synthesis framework based on a unified radiance transfer representation, addressing the challenging problem of colored lighting manipulation in the 3D-GAN setting. Our model learns the light transport functions under a white-light assumption from casual image datasets, without requiring expensive and inaccessible light-stage captured data. Since our model responds linearly to incident lighting and innately retains light transport consistency, it generalizes well to colored illumination and produces overall high-quality relighting results. The simple model design discards network evaluation for each light condition, adding real-time relightability for 3D head synthesis. Extensive evaluations demonstrate that our method achieves state-of-the-art 3D-aware relighting quality, superior real-time performance, and better spatial consistency compared with previous relighting methods.

\begin{figure}[t]
    \centering  \includegraphics[width=\linewidth]{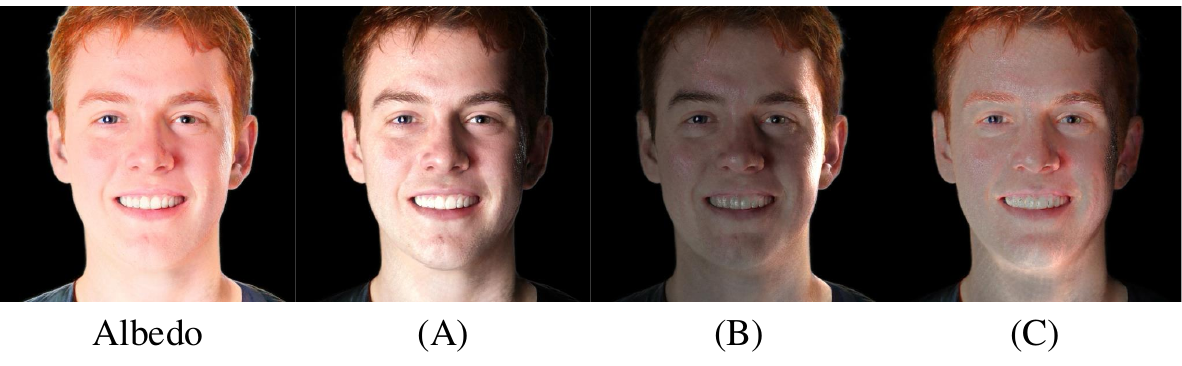}
    \caption{Failure cases. Albedo with highlight residuals and renderings under illumination from three directions are presented. Zoom in for a better view.}
    \label{fig::failure_case}   
\end{figure}

One limitation of our method is that it mainly models low-frequency and diffuse lighting effects due to the nature of radiance transfer in the spherical harmonics basis. Although the view-conditioned radiance transfer decoder models low-frequency view-dependent effects to some extent, our method is still not expressive enough for high-frequency reflections. In some cases, the model incorrectly learns the specular highlight in the albedo. As shown in Figure~\ref{fig::failure_case} (A), highlight on the right cheek remains in the albedo and leads to artifacts. In addition, relighting under illumination from below the face region may seem implausible, as shown in Figure~\ref{fig::failure_case} (C). This is because such illumination is scarce in in-the-wild images. 

For future work, it is interesting to model high-frequency and largely view-dependent effects on top of this explicit linear lighting model to support more complex lighting conditions other than spherical harmonics. Exploring the adaptation of our method to dynamic human head scenarios is also worth further investigation.
\end{sloppypar}

\begin{acks}
This work was supported by National Natural Science Foundation of China (No. 62322210), Beijing Municipal Science and Technology Commission (No. Z231100005923031), Innovation Funding of ICT, CAS (No. E461020), the Royal Society (IEC{\textbackslash}NSFC{\textbackslash}233596), Kuaishou Technology and CCF-Kuaishou Large Model Explorer Fund. We would like to thank Chunshuo Qiu for assisting with the demo video production and Zhongming Chen for assisting with figure making. We would also like to acknowledge the Nanjing Institute of InforSuperBahn OneAiNexus for providing the training and evaluation platform.

\end{acks}

\bibliographystyle{ACM-Reference-Format}
\bibliography{references}

\appendix

\end{document}